%% file: iclr2024_conference.tex
\documentclass{article} 
\usepackage{iclr2024_conference,times}

\input{math_commands.tex}

\usepackage{hyperref}
\usepackage{url}

\usepackage[frozencache,cachedir=.]{minted}

\usepackage{graphics} 
\usepackage{array}
\usepackage{ffcode}
\usepackage{epsfig} 
\usepackage{amsmath} 
\usepackage{amssymb}  

\usepackage{capt-of} 
\usepackage{cuted} 
\usepackage{multirow}
\usepackage{tablefootnote}
\usepackage{caption}
\usepackage{subcaption}
\usepackage{array} 
\usepackage{booktabs} %

\usepackage{ragged2e}

\usepackage{colortbl} 

\definecolor{lightgray}{gray}{0.9}
\definecolor{lightblue}{rgb}{0.93,0.95,1.0}
\definecolor{headingray}{gray}{0.85}
\definecolor{lightgreen}{rgb}{0.88,1.0,0.93}
\definecolor{lightred}{rgb}{0.975,0.91,0.91}
\definecolor{darkgray}{gray}{0.65}

\definecolor{lightgray}{gray}{0.9}

\usepackage{microtype}

\usepackage{listings}
\usepackage[T1]{fontenc}
\usepackage{latexsym}
\usepackage{dsfont}

\usepackage{makecell}
\usepackage[normalem]{ulem}
\usepackage{multicol}

\usepackage{csquotes}
\usepackage{paralist}
\usepackage{mdwlist}
\usepackage{array}
\usepackage{bm}
\usepackage{colortbl}
\usepackage{dsfont}
\usepackage{tabularx}
\usepackage{listings}
\usepackage{wrapfig}
\usepackage[ruled,vlined]{algorithm2e}

\title{ \includegraphics[trim={2cm 3cm 3cm 1cm}, width=0.65cm]{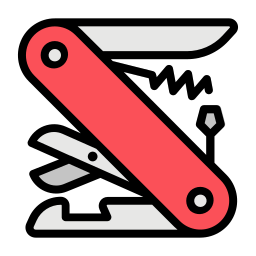} wissNYF: Tool Grounded LLM Agents for Black Box Setting}

\author{
    Somnath Sendhil Kumar\textsuperscript{1} \quad
    Dhruv Jain\textsuperscript{1} \quad
    Eshaan Agarwal\textsuperscript{1} \quad 
    Raunak Pandey\textsuperscript{1} \quad \vspace{5pt} \\
    \textsuperscript{1} \href{https://cops-iitbhu.github.io/IG-website/}{Intelligence Group, IIT (BHU), Varanasi} \quad 
}

\iclrfinalcopy 
\begin{document}


\maketitle

\begin{abstract}
While Large Language Models (LLMs) have demonstrated enhanced capabilities in function-calling, these advancements primarily rely on accessing the functions' responses. This methodology is practical for simpler APIs but faces scalability issues with irreversible APIs that significantly impact the system, such as a database deletion API. Similarly, processes requiring extensive time for each API call and those necessitating forward planning, like automated action pipelines, present complex challenges. Furthermore, scenarios often arise where a generalized approach is needed because algorithms lack direct access to the specific implementations of these functions or secrets to use them. Traditional tool planning methods are inadequate in these cases, compelling the need to operate within black-box environments. Unlike their performance in tool manipulation, LLMs excel in black-box tasks, such as program synthesis. Therefore, we harness the program synthesis capabilities of LLMs to strategize tool usage in black-box settings, ensuring solutions are verified prior to implementation. We introduce \textbf{TOPGUN}, an ingeniously crafted approach leveraging program synthesis for black box tool planning. Accompanied by \textbf{SwissNYF}, a comprehensive suite that integrates black-box algorithms for planning and verification tasks, addressing the aforementioned challenges and enhancing the versatility and effectiveness of LLMs in complex API interactions. The public code for SwissNYF is available at \ \href{https://github.com/iclr-dummy-user/SwissNYF}{https://github.com/iclr-dummy-user/SwissNYF}
\end{abstract}

\section{Introduction}
\label{sec:intro}

\begin{figure*}[!bp]
\centering
\includegraphics[width=0.9\linewidth]{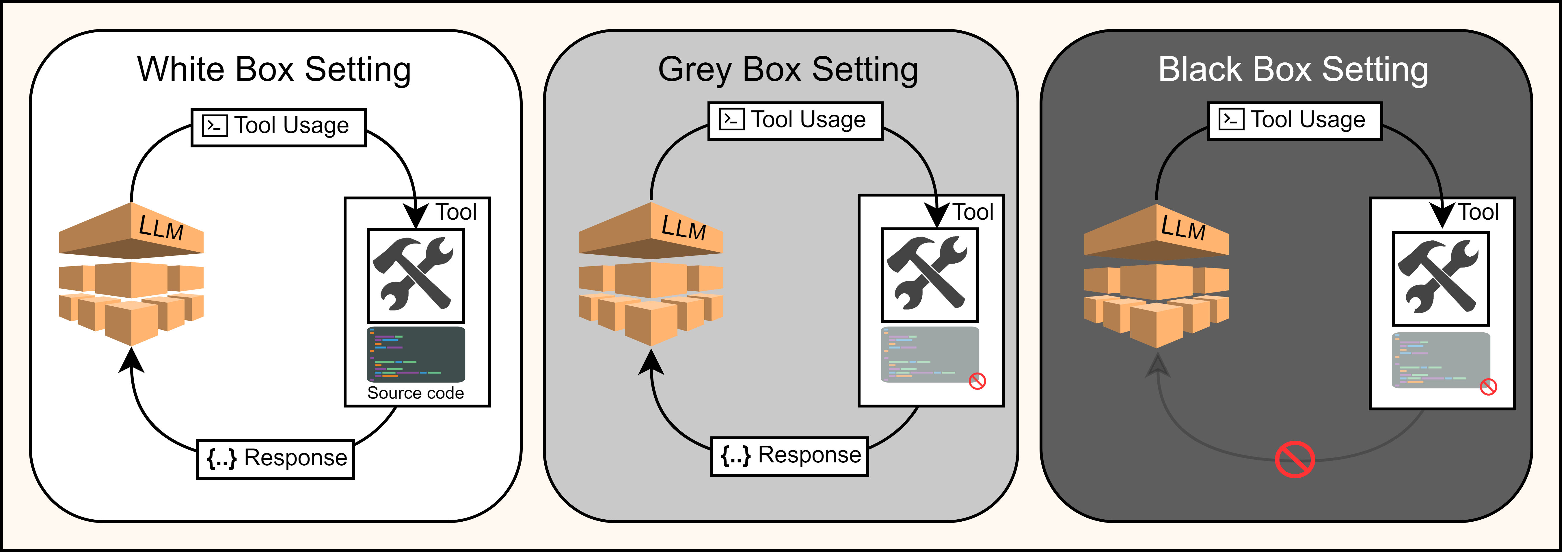}
\caption{\label{fig:blackbox}Illustration of different settings that an LLMs may require to manipulate tools.}
\end{figure*}

Significant advancements in Large Language Models (LLMs) like GPT (\cite{radford2018improving}; \cite{radford2019language}; \cite{brown2020language}; \cite{achiam2023gpt}) and PaLM (\cite{chowdhery2023palm};\cite{anil2023palm};) have demonstrated profound abilities in reasoning and following instructions over an extensive array of tasks \cite{huang-chang-2023-towards}. The recent shift towards leveraging LLMs to interact with external tools for addressing complex real-world challenges marks a significant area of interest (\cite{hao2023toolkengpt}; \cite{zhang2023evaluating}; \cite{zhuang2023toolqa}; \cite{yang2023gpt4tools}; \cite{schick2023toolformer};\cite{lu2023chameleon};). In addressing intricate problems, autonomous agents powered by LLMs employ an amalgamation of LLMs and various external tools (APIs), crafting solutions that necessitate a sequence of intermediate reasoning steps (\cite{schick2023toolformer};\cite{lu2023chameleon};\cite{lu2023chameleon};\cite{patil2023gorilla};\cite{qin2023toolllm}). When presented with a problem,  These agents' primary objective is to identify and execute a series of API function calls sequentially, leading to a coherent solution. These approaches are ineffective when queries lack transparency or when the APIs are irreversible.


We coin the term "black-box" settings in the context of tool planning as scenarios where the outcomes of an API or tool are not observable. This framework is especially pertinent in systems where using certain APIs poses risks, such as those causing inconsistencies by deleting or updating database entries, canceling jobs, or performing similar operations. It's also relevant where API experimentation incurs high costs or when APIs require considerable time to execute, ensuring clarity and comprehensive coverage without redundancy, making it challenging to interpret their outcomes. We present a taxonomy of such systems Fig. \ref{fig:blackbox} into three branches: 
\begin{enumerate}
    \item \textbf{White Box Systems}: In these settings, planners can invoke the API, receive responses, access the source code and understand its complex logic. This access enables the system to navigate complex inputs, intricacies and use cases efficiently.
    \item \textbf{Gray Box Systems}: Planners in these environments have descriptions of the tools at their disposal and the capability to call the API and receive responses. The system's planning relies solely on the limited descriptions provided and the responses for each tool.
    \item \textbf{Black Box Systems}: In the most challenging scenarios, planners are confined to tool descriptions without access to actual tool outputs. Here, the planner must decipher the dynamics of each tool based solely on its description, making it a particularly demanding task to formulate responses to queries.
\end{enumerate}

The \cite{zhuang2023toolchain} and \cite{qin2023toolllm} methods excel in straightforward scenarios where an agent can iterate over tools to identify the optimal path, yet they lack efficiency and necessitate extensive exploration. Approaches like \cite{yao2022react} and \cite{parisi2022talm}, subsets of this exploratory paradigm, offer enhanced efficiency yet frequently falter due to their constrained directionality in tool search, making them suitable predominantly for straightforward API challenges. In contrast, the \cite{zhang2023reverse} approach is efficient regarding API execution costs by constraining the number of calls. However, it omits any form of verification for its proposed trajectory, diminishing its precision in practical applications. 

These methodologies in tool application present a dichotomy between accuracy and computational overhead. While generally unsuitable for black-box settings, the Reverse chain approach exhibits potential for adaptation within such frameworks. On the other hand, program synthesis-based algorithms have been instrumental in exalting reasoning and decision-making capabilities within LLMs, offering a more naturally associative decision-making process than that afforded by mere text.
Works like The Chain of Code \cite{li2023chain} and Program-of-thoughts \cite{chen2022program} are great examples of using code generation to improve decision-making for answering general open-domain questions. 
To this end, few works also upheld the reasoning capability of LLMs using code like "TORA: A Tool-Integrated Reasoning Agent for Mathematical Problem Solving" \cite{gou2023tora}, "Solving challenging math word problems using gpt-4 code interpreter with code-based self-verification" \cite{zhou2023solving} and "PAL: Program-aided Language Models" \cite{gao2023pal} have exploited code interpreters for zero-shot verified solving, substantially surpassing few-shot learning benchmarks by enabling semi-verification of proposed solutions.

However, works like \cite{paranjape2023art}, which employs code synthesis for tool usage, are restricted by their limited toolset and the scalability challenge posed by the need for extensive human feedback and interventions and the need for the human expert to be familiar with the whole toolset. Similarly, works such as \cite{xu2023lemur}, which deploys language models for real-time code generation and command execution within controlled environments, are limited by their narrow tool range and a deficit in generalizability. The state-of-the-art approaches on HumanEval \cite{chen2021evaluating} and HumanEval-X \cite{zheng2023codegeex} datasets for code generation, like Reflexion \cite{shinn2023reflexion} and LATS \cite{latszhou2023language}, which iterate upon code based on interpreter outputs and reflect over them, these approaches have yet to be experimented with in other domains associated with LLMs.

To bridge these gaps, we introduce the \textbf{TOPGUN} (\textbf{T}ool \textbf{O}rchestration and \textbf{P}rogram synthesis for \textbf{G}eneralizing over \textbf{UN}known systems) framework, which unifies code generation, reasoning, and strategic tool planning designed for complex tasks. TOPGUN also verifies the execution plans and does so with exceptional efficiency in API cost, effectively addressing the limitations of preceding models.

Key contributions of our work are summarized as follows:
\begin{enumerate}
    \item To the best of our knowledge, we are the First to coin the term Black Box setting for API usage and developed a suite to encourage the development of algorithms for such scenarios.
    \item We leverage the program synthesis capabilities of Large Language Models (LLMs) to augment their efficacy in tool usage substantially, showcasing a notable enhancement in performance.
    \item We present a robust and cost-efficient framework for scalable solutions across a wide array of open-domain queries, even when faced with limited knowledge of user data/tools. It is also publically hosted to demonstrate the same.\footnote{\href{https://swiss-nyf.azurewebsites.net/}{https://swiss-nyf.azurewebsites.net/}}
\end{enumerate}

This paper details our methodology and its evaluation by first elucidating the background on Tool planning \ref{subsec:probform} and Code generation using LLM \ref{subsec:codegen} followed by detailing individual components of the pipeline \ref{sec:swissnyf}. Our evaluation is bifurcated into two segments: initially, we undertake a gray box \ref{subsec:grayboxeval} across principal datasets, and subsequently, we delve into a black box setting \ref{subsec:blackboxeval}. For the latter, we have curated a bespoke dataset employing Toolbench prompts, intentionally adjusting the dataset to include only limited documentation of widely used libraries. This adjustment aims to validate the generalizability of our approach. Additionally, we juxtapose our methodology with a tailored variant of the Reverse Chain method to scrutinize performance disparities.

\section{Preliminaries}
\label{sec:prelim}
\subsection{Problem Formulation}
\label{subsec:probform}

\begin{figure}[!tbp]
  \centering
  {\includegraphics[width=1\textwidth]{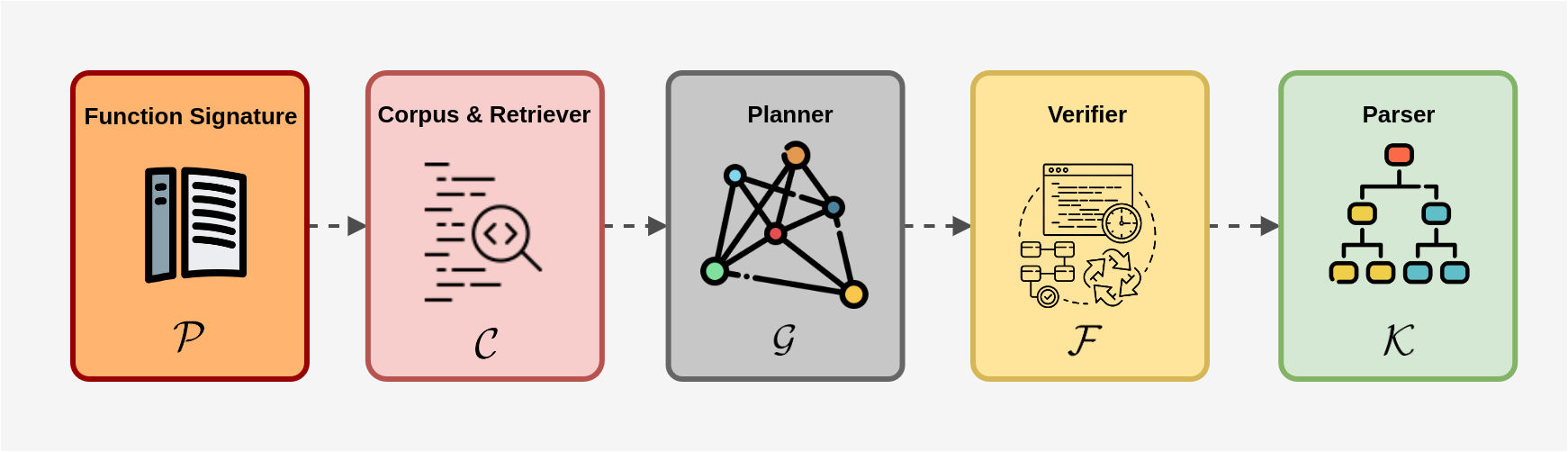}}
  \caption{\label{fig:pipeline}Illustration of SwissNYF pipeline for tool usage in Black Box setting.}
  \vspace{-3ex}
\end{figure}

Tool planning within the context of a Large Language Model (LLM), denoted as \(\rho\), involves leveraging a selection of tools from a pool of \(n\) candidate tools in the corpus \(\mathcal{C}\), represented as \(\mathcal{C} = \{t_0, t_1, \ldots, t_n\}\), to effectively address a user's query \(q\). The primary goal is to formulate a meticulous plan, known as the Solution Trajectory \(St\), for the orchestration of these tools. The Solution Trajectory \(St\), which outlines the sequential execution of tools, is crafted to directly address the query \(q\). The LLM agent, or planner \(\mathcal{G}\), is responsible for planning or generating \(St\) from \(\mathcal{C}\), formalized as \(St \leftarrow \mathcal{G}(q, \rho, \mathcal{C})\). This process ensures a structured and coherent response strategy, aligning the tools' capabilities with the query's specific requirements for an effective solution.

\subsection{Code Generation}
\label{subsec:codegen}


The integration of Reflexion \cite{shinn2023reflexion} with Large Language Models (LLM) \(\rho\) and Python Interpreter \(\mathcal{I}\) has significantly advanced coding tasks by enabling iterative code refinement. This approach leverages feedback \(\mathcal{F}\) to iteratively address exceptions and enhance initial code output \(c\), guided by test cases dynamically generated by \(\rho\) itself. This ensures comprehensive verification and refinement within a Function Call module, leading to a finalized code \(c_n\). This methodology enhances code quality and aligns with contemporary standards, marking a leap in automated code development and verification. This process of iterative code generation can be mathematically denoted as Eq. \ref{eq:reflexionprelim}
\begin{equation}\label{eq:reflexionprelim}
    \begin{gathered}
    c_{i} \leftarrow \rho(q, \textit{feedback}_{i-1}, c_{i-1}) \\
    \textit{output} \leftarrow \mathcal{I}(c_{i}) \\
    \textit{feedback}_{i}, \textit{verified} \leftarrow \mathcal{F}(output) \\
    \end{gathered}
\end{equation}

\section{SwissNYF}
\label{sec:swissnyf}
\subsection{Overview}
\label{subsec:swissoverview}

\begin{figure}[!tbp]
  \centering
  {\includegraphics[width=1\textwidth]{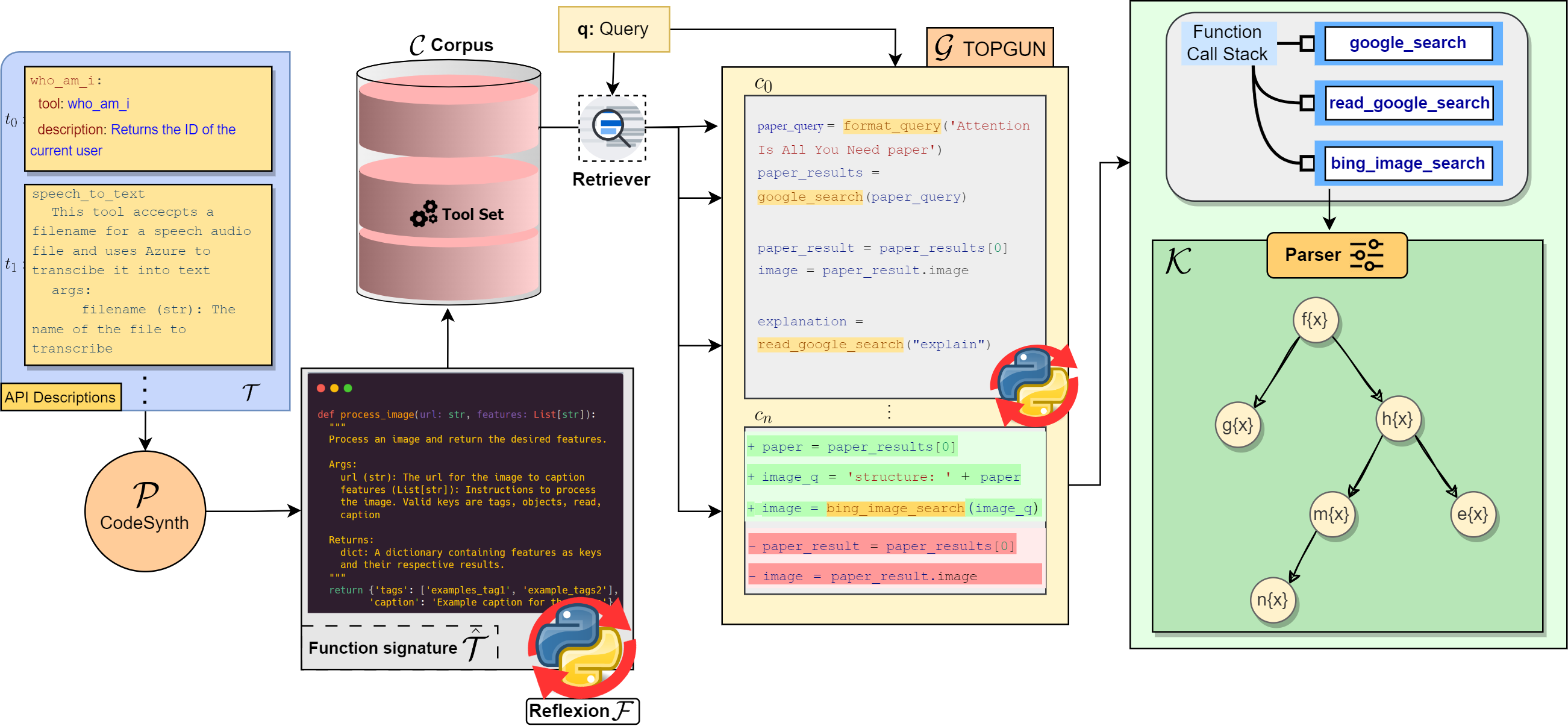}}
  \caption{\label{fig:topgunpipe}Detailed pipeline of our proposed approach with TOPGUN in SwissNYF}
  \vspace{-2ex}
\end{figure}

In this section, we introduce SwissNYF, a suite that enables LLM-based agents to efficiently navigate the action space to identify a valid solution for problem-solving in a black box scenario. SwissNYF is composed of five major components i.e., Function Signature Generation \(\mathcal{P}\), Corpus \& Retriever \(\mathcal{C}\), Planner \(\mathcal{G}\), Verifier \(\mathcal{F}\) and Parser \(\mathcal{K}\) as in Fig. \ref{fig:pipeline}. We explain individual components of the pipeline in the subsequent subsections.

\begin{figure*}[!tbp]
  \begin{subfigure}[b]{0.48\textwidth}
  \includegraphics[trim= 0.0cm 0.0cm 0.0cm 0.0cm, clip,  width=1.\linewidth]{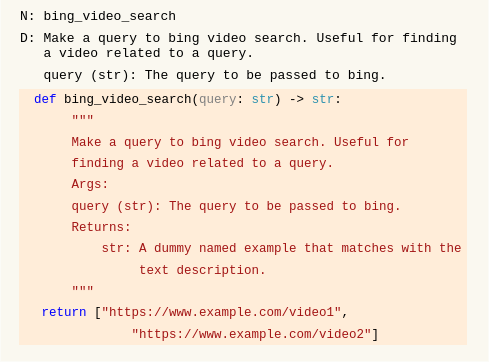}
  \caption{\label{fig:codesynthblock} Example output of CodeSynth \(\mathcal{P}\) Algorithm }
  \end{subfigure}
  \begin{subfigure}[b]{0.48\textwidth}
  \includegraphics[trim= 0.0cm 0.0cm 0.0cm 0.0cm, clip,  width=1.\linewidth]{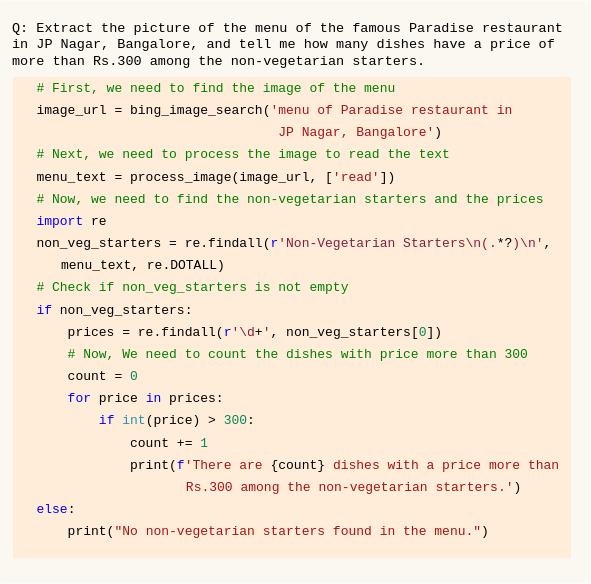}
  \caption{\label{fig:topgunblock} Example output of TOPGUN \(\mathcal{G}\) Algorithm}
  \end{subfigure}
  \caption{Illustration of pseudo function and tool planning generated by CodeSynth and TOPGUN, respectively.}
  \vspace{-2ex}
\end{figure*}

\subsection{Function Signature Generation}
\label{subsec:funcsign}


Function signatures, conceptualized as pseudo APIs, serve to emulate the behaviour of real API functions based on given tool descriptions. This emulation is crucial for two primary reasons in our tool planning methodology: firstly, they act as stand-ins for actual API calls, thereby enabling LLMs to plan and execute tasks with higher efficiency; secondly, they are treated as pre-defined functions, facilitating the transformation of tool augmentation into a task akin to code generation, using these pseudo functions. These function signatures are distinguished by their docstrings and an example return object that aligns with the tool description, equipping the planner with the necessary means to effectively address user queries. In the context of our SwissNYF implementation, we have adopted a straightforward yet effective method for generating these function signatures, termed \textbf{CodeSynth}. The efficacy of this approach is further analyzed in \ref{subsec:codesyntheval}.

\subsubsection{CodeSynth}
\label{subsec:codesynth}

For a given set of tool descriptions \(t \in \mathcal{T}\), we direct the Large Language Model (LLM) \(\rho\) to generate pseudo-function implementations, denoted as \(\hat{t}\). Our primary objective is to ensure that the arguments and return types of these pseudo-functions remain consistent with their descriptions. Additionally, we craft detailed docstrings for each pseudo-function to facilitate subsequent processes. A critical aspect of CodeSynth is the inclusion of an example return value, which is designed to mimic all potential operations the returned object might undergo during the verification process. The output generated by CodeSynth is illustrated in Fig. \ref{fig:codesynthblock}. Moreover, the code generation facilitated by this block benefits from validation through Reflexion, as outlined in Eq. \ref{eq:reflexionprelim}. Ultimately, the methodologies applied within CodeSynth can be encapsulated in Algo. \ref{alg:codesynth}.

\begin{wrapfigure}{r}{0.5\linewidth}
  \vspace{-5ex}
\begin{algorithm}[H]

  \begin{small}
    \KwIn{$\rho$: large language model; $T$: tool descriptions; $\mathcal{I}$: python interpreter; $\mathcal{F}(\mathcal{I})$: reflexion feedback of $\mathcal{I}$; $\mathcal{C}$: empty corpus of pseudo tools} 
    
    \For{\(t = 1, 2, \cdots, T\)}{
    {
        \(\hat{t}_{0} \leftarrow \rho(t)\) \quad\quad \quad \quad \quad \quad \quad \ \ \textcolor{blue}{//\textit{Pseudo code}} \\
        \(verified \leftarrow \mathcal{I}(\hat{t}_0)\) \\ 
        
        \While{not verified}{
        {
            \(\hat{t}_{i} \leftarrow \rho(t, feedback_{i-1}, \hat{t}_{i-1})\) \\
            \(feedback_i, \    verified \leftarrow \mathcal{F}(\mathcal{I}( \hat{t}_{i})) \) \\
        }
        }
        Update \(\mathcal{C} \leftarrow \hat{t}_{n}\) \quad  \quad \quad \quad \quad  \textcolor{blue}{// \textit{Update Corpus}}\\
    }
    }
    \KwOut{A corpus of verified psuedo functions \(\mathcal{C}\)}
  \end{small}
  \caption{ \(\mathcal{P}\): CodeSynth }
  \label{alg:codesynth}
\end{algorithm}
  \vspace{-6ex}

\end{wrapfigure}

Utilizing the Function Calling module alongside the Interpreter, we rigorously test the pseudo-functions against a wide range of real-world scenarios. This approach guarantees that the test cases are comprehensive and reflective of actual function usage, allowing us to gather detailed feedback on the pseudo-functions' performance. Such feedback is vital for the iterative improvement of the pseudo-functions, significantly enhancing their reliability and applicability in practical settings. Prompts for CodeSynth can be documented in \ref{A:promptcodesynth}.

\subsection{Corpus and Retriever}
\label{subsec:retriever}


The function signatures, crucial components of our methodology, are systematically stored within a corpus for future utilization by any planning system. This corpus facilitates the indexing of tool descriptions, enabling the precise retrieval of the most appropriate tool based on the index. Notably, the literature documents several advanced retrieval systems designed for this purpose, demonstrating exceptional accuracy. These include ToolBench IR \cite{qin2023toolllm}, APIRetriever \cite{zan2022private}, Instructor-XL \cite{su2022one}, and GEAR \cite{lu2023gear}. Our framework incorporates these retrievers, with Instructor-XL set as the default option, owing to its proven efficacy. Furthermore, we are actively exploring the integration of AnyTool's Hierarchical API Retriever \cite{du2024anytool}, anticipating significant enhancements to our tool retrieval capabilities. This strategic inclusion of multiple retrievers ensures our system remains versatile and effective in identifying the most suitable tools for a given task, aligning with the latest advancements in retrieval technology.

\subsection{Planner}
\label{subsec:planner}


We have implemented two planning approaches in our framework. The first leverages a modified Reverse Chain \cite{zhang2023reverse} to support multiple end function calls by decomposing tasks into subtasks and creating sub-trees with the original reverse chain technique. The second, \textbf{TOPGUN}, is our proposed code-driven planning algorithm, designed for speed, efficiency, consistency, and accuracy, especially in black box scenarios. \textbf{TOPGUN} offers a streamlined alternative to traditional planning methods, optimizing for complex system navigation and task execution with greater reliability and cost-effectiveness.


\subsubsection{TOPGUN}
\label{subsec:topgun}

\begin{wrapfigure}{r}{0.51\linewidth}
\vspace{-3ex}
\begin{algorithm}[H]
  \begin{small}
	\KwIn{q: query; $\rho$: large language model; $T$: tool descriptions; $\mathcal{I}$: python interpreter; $\mathcal{F}(\mathcal{I})$: reflexion feedback of $\mathcal{I}$; $\mathcal{C}$: empty corpus of pseudo tools; $\mathcal{P}$: Codesynth, $\mathcal{K}$: parser} 
 
    Initialize \(\mathcal{\hat{T}} \leftarrow \mathcal{P}(\rho, T, \mathcal{I}, \mathcal{F}, \mathcal{C})\) \quad   \textcolor{blue}{// \textit{Pseudo tools}}\\
    \(c_0 \leftarrow\  \rho(q, \mathcal{\hat{T}}, \mathcal{C})\) \quad \quad \quad \quad \quad \quad \quad \textcolor{blue}{// \textit{Code for query}} \\
    \(verified \leftarrow \mathcal{I}(c_0, \mathcal{\hat{T}})\) \quad \textcolor{blue}{// \textit{Verify with pseudo tools}}
    
    \While{not verified}{
    {
        \(c_i \leftarrow \ \rho(q, \mathcal{\hat{T}}, \mathcal{\hat{C}}, feedback_{i-1}, c_{i-1})\) \\
        \(feedback_i, \    verified \leftarrow \mathcal{F}(\mathcal{I}(c_{i}, \mathcal{\hat{T}}))\)\\
    }
    }
    \(\mathcal{S}t \leftarrow \mathcal{K}(c_{n})\) \quad \quad \quad \quad \quad \quad \quad  \textcolor{blue}{// \textit{Solution Trajectory}}\\
 
    \KwOut{A solution trajectory \(\mathcal{S}t\) and \(c_n\) code for execution and evaluation}
  \end{small}
  \caption{ \(\mathcal{G}\): TOPGUN }
  
  \label{alg:topgun}
  
\end{algorithm}
    \vspace{-2ex}
\end{wrapfigure}

\textbf{TOPGUN}, an acronym for \textbf{T}ool \textbf{O}rchestration and \textbf{P}rogram synthesis for \textbf{G}eneralizing over \textbf{UN}known systems, redefines the approach to addressing user queries \(q\) by framing the challenge as a task of code generation. Utilizing pseudo-functions \(\mathcal{\hat{T}}\) as functions available to TOPGUN enables the agent to construct an accurate sequence of function calls \(c_0 \leftarrow \rho(q, \mathcal{\hat{T}}, \mathcal{C})\), effectively depicted in Fig. \ref{fig:topgunblock}. Leveraging Reflexion detailed in Eq.\ref{eq:reflexionprelim}, the framework iteratively refines responses to the query. The synthesis of these components into the comprehensive algorithm is presented in Algo. \ref{alg:topgun} showcases TOPGUN's capability to navigate through various solution paths. Unlike traditional traversal-based techniques, TOPGUN capitalizes on the inherent code-generation capabilities of LLMs, facilitating a more direct and efficient solution process. This distinction not only enhances efficacy by pinpointing issues with precision but also ensures adaptability in black box scenarios, simultaneously optimizing performance in gray box settings. A detailed pipeline overview with TOPGUN in place is given in Fig.\ref{fig:topgunblock}. With prompts documented in \ref{A:prompttopgun}.

\begin{figure}
    \centering
    \includegraphics[width=0.97\linewidth]{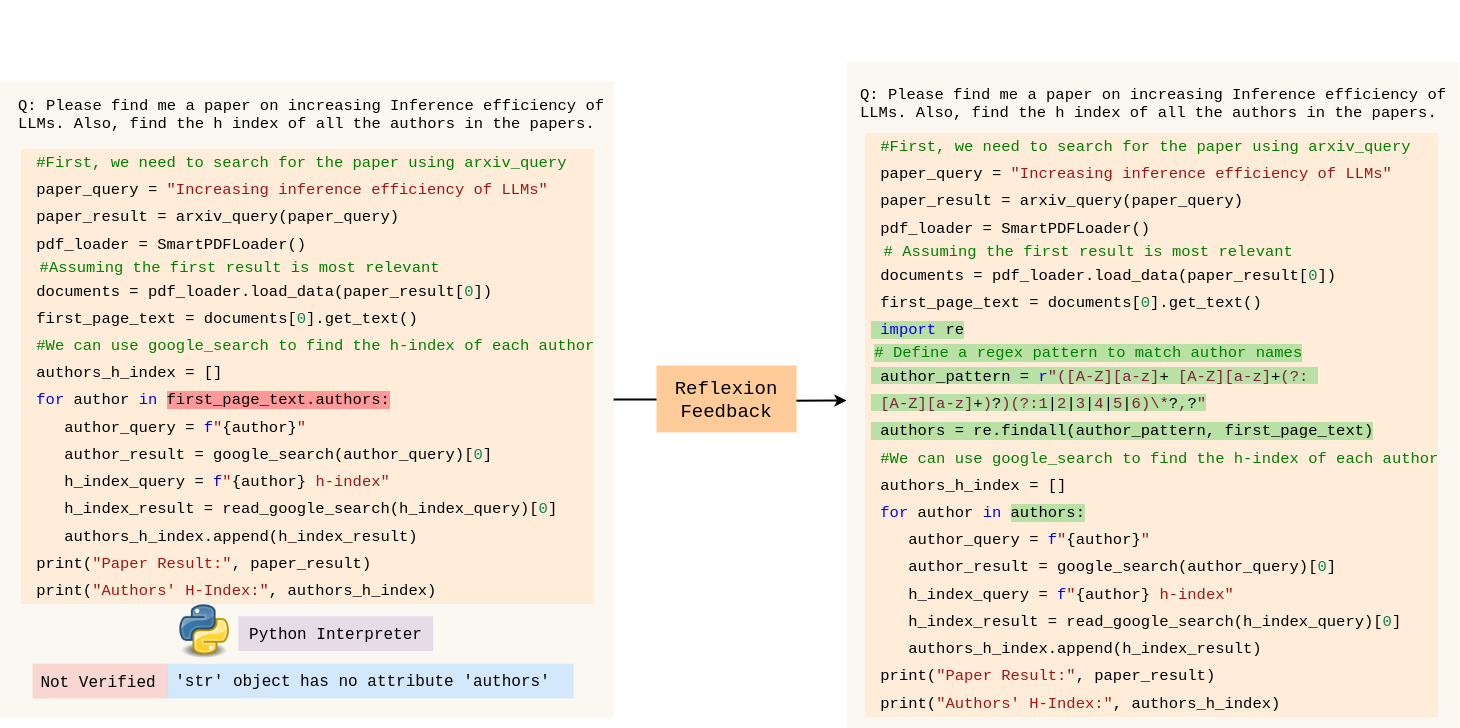}
    \caption{Illustartion of Self-Reflection Mechanism in TOPGUN}
    \label{fig:reflexion}
    \vspace{-2ex}
\end{figure}

\subsection{Verifier}
\label{subsec:verifier}


Verification is closely linked to the functionality of the Planner \(\mathcal{G}\), relying on both the nature of \(\mathcal{G}\)'s output and its ability to incorporate feedback. Although verification initially serves as a preparatory step prior to parsing, it also plays a crucial role in refining outputs by providing feedback that \(\mathcal{G}\) can use for subsequent iterations. 

In our framework, we leverage Reflexion \cite{shinn2023reflexion}, detailed in Eq. \ref{eq:reflexionprelim} and depicted in Algo. \ref{alg:topgun}, to seamlessly integrate verification and feedback within the TOPGUN methodology. This eliminates the requirement for an additional function call module, concentrating instead on directly executing code pertinent to the user query. 
This approach is illustrated in Fig. \ref{fig:reflexion}, providing a visual representation of the concept.


\subsection{Parser}
\label{subsec:parser}

The Parser \(\mathcal{K}\), akin to the Verifier \(\mathcal{F}\), is intrinsically dependent on the Planner \(\mathcal{G}\) for its functionality. Its pivotal output is a well-defined Solution Trajectory \(St\), mapping out the sequence of tool applications devised to address the query. In employing the Reverse Chain technique, our methodology involves synthesizing individual sub-trees into a singular, comprehensive tree through the capabilities of LLM \(\rho\). 
The process's efficacy is markedly improved by the judicious reuse of elements from the individual trees during their amalgamation.

Conversely, for the \textbf{TOPGUN} methodology, we adopt the established Abstract Syntax Tree (AST) paradigm \cite{fischer2007abstractsyntaxtree} to segment the program into fundamental function calls, alongside specifying their arguments and return values. This segmentation is instrumental in constructing a systematic series of tool invocations. This meticulously arranged series, denoted as \(St\), is succinctly formalized as \(St \leftarrow \mathcal{K}(c_n)\). 

The entire pipeline, as depicted in Figure \ref{fig:topgunpipe}, emerges from the integration of various components designed to effectively address user queries through the strategic orchestration of tools within the SwissNYF framework.

\begin{table}[htbp]
\centering
\caption{Win Rate of different Candidate and Reference model over G1 set}
\label{tab:winrate1}
\begin{tabular}{@{}ccccc@{}}
\toprule
\rowcolor{headingray}Candidate & Reference & G1-Instruction & G1-Tool & G1-Category  \\ 
\midrule
\cellcolor{lightgray}T.LLaMA ReACT & \cellcolor{lightred}ChatGPT ReACT & 45.0 & 42.0 & 47.5  \\ 
\cellcolor{lightgray}T.LLaMA DFSDT & \cellcolor{lightred}ChatGPT ReACT & 55.0 & 55.3 & 54.5  \\ 
\cellcolor{lightgray}T.LLaMA DFSDT+Ret & \cellcolor{lightred}ChatGPT ReACT & 62.3 & 59.0 & 55.0 \\ 
\cellcolor{lightgray}ChatGPT DFSDT & \cellcolor{lightred}ChatGPT ReACT & 60.5 & 62.0 & 57.3  \\ 
\cellcolor{lightgray}GPT4 ReACT & \cellcolor{lightred}ChatGPT ReACT & 60.0 & 58.8 & 63.5  \\ 
\cellcolor{lightgray}GPT4 DFSDT & \cellcolor{lightred}ChatGPT ReACT & 67.5 & 67.8 & 66.5 \\ 
\cellcolor{lightgray}GPT4 TOPGUN & \cellcolor{lightred}ChatGPT ReACT & \textbf{88.192} & \textbf{87.46} & \textbf{87.15}  \\ 
\midrule
\cellcolor{lightblue}GPT4 TOPGUN & \cellcolor{lightred}ChatGPT DFSDT & 78.49 & 77.55 & 76.24  \\ 
\cellcolor{lightblue}GPT4 TOPGUN & \cellcolor{lightred}T.LLaMA ReACT & 86.72 & 82.94 & 80.80  \\ 
\cellcolor{lightblue}GPT4 TOPGUN & \cellcolor{lightred}T.LLaMA DFSDT & 81.75 & 75.51 & 73.81  \\ 
\cellcolor{lightblue}GPT4 TOPGUN & \cellcolor{lightred}T.LLaMA DFSDT+Ret & 80.35 & 77.11 & 75.39  \\ 
\midrule
\cellcolor{lightgreen}GPT4 TOPGUN & \cellcolor{lightred}GPT4 ReACT & 82.996 & 79.956 & 77.633 \\ 
\cellcolor{lightgreen}GPT4 TOPGUN & \cellcolor{lightred}GPT4 DFSDT & \textbf{82.065} & \textbf{73.69} & \textbf{71.14} \\ 

\bottomrule
\end{tabular}
\end{table}

\section{Experiments}
\label{sec:experiments}


Tool planning datasets, while diverse, often fall short in supporting multi-turn and multi-call dialogues, as seen in works by \cite{schick2023toolformer} and \cite{tang2023toolalpaca}, and lack precise evaluation metrics, complicating thorough assessments. Even comprehensive datasets like ToolBench by \cite{qin2023toolllm} struggle with aligning to black-box settings, presenting significant challenges for evaluating tool planning in such scenarios.

Our evaluation employs the ToolBench benchmark \cite{qin2023toolllm} and a specially curated dataset for unchar codebases, assessed in both gray (\ref{subsec:grayboxeval}) and black box (\ref{subsec:blackboxeval}) settings. We benchmark our TOPGUN approach against existing methods using win rate, token count, and success rate. Additionally, we scrutinize CodeSynth's (\(\mathcal{P}\)) impact on the Planner's (\(\mathcal{G}\)) performance and independently evaluate its ability to generate effective function signatures, acting as pseudo functions, detailed in Section \ref{subsec:codesyntheval}.

\begin{table}[htbp]
\centering
\caption{Win Rate of different Candidate and Reference model over G2, G3 set and Average over all sets}
\label{tab:winrate2}
\hspace*{-3.1em}
\begin{tabular}{@{}cccccc@{}}
\toprule
\rowcolor{headingray}Candidate & Reference & G2-Instruction & G2-Category & G3-Instruction & Average \\ 
\midrule
\cellcolor{lightgray}T.LLaMA ReACT & \cellcolor{lightred}ChatGPT ReACT & 50.8 & 41.8 & 55.0 & 47.0 \\ 
\cellcolor{lightgray}T.LLaMA DFSDT & \cellcolor{lightred}ChatGPT ReACT & 68.5 & 58.0 & 69.0 & 60.0 \\ 
\cellcolor{lightgray}T.LLaMA DFSDT+Ret & \cellcolor{lightred}ChatGPT ReACT & 68.5 & 60.8 & 73.0 & 63.1 \\ 
\cellcolor{lightgray}ChatGPT DFSDT & \cellcolor{lightred}ChatGPT ReACT & 72.0 & 64.8 & 69.0 & 64.3 \\ 
\cellcolor{lightgray}GPT4 ReACT & \cellcolor{lightred}ChatGPT ReACT & 65.8 & 60.3 & 78.0 & 64.0 \\ 
\cellcolor{lightgray}GPT4 DFSDT & \cellcolor{lightred}ChatGPT ReACT  & 73.3 & 63.3 & 84.0 & 70.4 \\ 
\cellcolor{lightgray}GPT4 TOPGUN & \cellcolor{lightred}ChatGPT ReACT & \textbf{87.59} & \textbf{78.78} & \textbf{90.05} & \textbf{86.54} \\ 
\midrule
\cellcolor{lightblue}GPT4 TOPGUN & \cellcolor{lightred}ChatGPT DFSDT & 81.63 & 73.07 & 85.26 & 78.71 \\ 
\cellcolor{lightblue}GPT4 TOPGUN & \cellcolor{lightred}T.LLaMA ReACT & 86.24 & 77.71 & 93.23 & 84.61 \\ 
\cellcolor{lightblue}GPT4 TOPGUN & \cellcolor{lightred}T.LLaMA DFSDT & 78.31 & 71.80 & 89.47 & 78.44 \\ 
\cellcolor{lightblue}GPT4 TOPGUN & \cellcolor{lightred}T.LLaMA DFSDT+Ret & 83.07 & 72.92 & 87.82 & 79.44 \\ 
\midrule
\cellcolor{lightgreen}GPT4 TOPGUN & \cellcolor{lightred}GPT4 ReACT & 78.61 & 73.75 & 93.68 & \textbf{80.27}\\ 
\cellcolor{lightgreen}GPT4 TOPGUN & \cellcolor{lightred}GPT4 DFSDT & \textbf{73.92} & \textbf{71.35} & \textbf{79.25} & \textbf{78.59} \\ 
\bottomrule
\end{tabular}
\end{table}

\subsection{Gray Box Evaluation}
\label{subsec:grayboxeval}


To assess the performance of \textbf{TOPGUN} and compare it with other gray box methodologies such as ReACT and DFSDT, we maintain the integrity of our pipeline while adapting the evaluation process to incorporate actual functions in place of pseudo functions within the output solution trajectory. This approach effectively leaves our black box pipeline intact while converting it into a gray box evaluation framework. The necessity of responses and Final answers for evaluation purposes has led us to adopt this hybrid strategy. In practical scenarios, this mirrors the process where a generalist planner delivers a strategy to the client, who then substitutes pseudo-function implementations with their real functions for execution. For this evaluation, we employ ToolBench, as detailed by \cite{qin2023toolllm}, and conduct our analysis across all problem categories provided in the dataset. Further elaboration on the precise evaluation methodology and the application of ToolBench is documented in \ref{subsec:toolbenchgreyeval}.


\textbf{Results :} Win Rate comparisons for ToolLLaMa-ReACT, ToolLLaMA-DFSDT, ChatGPT-DFSDT, GPT4-DFSDT, and GPT4-TOPGUN against ChatGPT-ReACT and GPT4-TOPGUN are summarized, with averages taken from 7 runs per model pair, detailed in Tables \ref{tab:winrate1} and \ref{tab:winrate2}. TOPGUN significantly surpassed ReAct and DFSDT in all categories, achieving win rates of \textbf{80.27\%} versus GPT4-ReACT, \textbf{78.59\%} against GPT4-DFSDT, and \textbf{86.54\%} against ChatGPT-ReACT, showing improvements of \textbf{22.54\%} and \textbf{16.14\%} respectively. These results highlight TOPGUN's superior ability to create tool plans that align with preference evaluation criteria across various conditions.

\subsection{Black Box Evaluation}
\label{subsec:blackboxeval}

\begin{wrapfigure}{l}{0.54\textwidth}
    \vspace{-1ex}
    \centering
    \includegraphics[width=1\linewidth, trim={0.7cm 1cm 2cm 2cm}, clip]{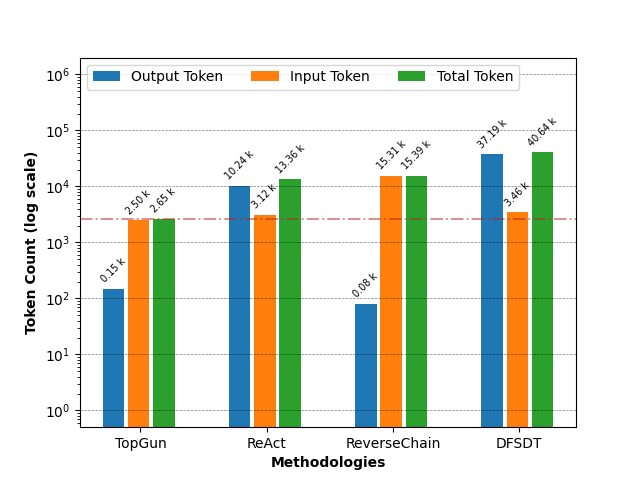}
    \caption{\label{fig:token_count}Average Token Consumption of individual methodologies in Black Box setting.}
    \vspace{-2ex}
\end{wrapfigure}

Utilizing the Data Generation pipeline from \cite{qin2023toolllm}, we constructed a black-box scenario dataset featuring 36 LLaMa-Hub \cite{LLamaHub} tools and unique functions from private libraries. Following \cite{zan2022private}, we converted Pandas and Numpy into Monkey and BeatNum packages, renaming all internal functions and structures to test planner generalizability without LLM prior knowledge. This dataset, detailed at \ref{A:promptdatagenprivate}, focuses on accuracy of the solution trajectory, with each query designed for a single correct path. After manual annotation, it comprises 100 queries and 162 tools, with samples and TOPGUN outcomes at \ref{A:dataexampleprivate} and \ref{A:topgunprivateexample}.

\textbf{Results :} The black-box evaluation, featuring TOPGUN and a revised Reverse Chain, utilizes \(\mathcal{P}\) function signatures for a comprehensive black-box methodology. TOPGUN surpasses Reverse Chain and undergoes comparison with GPT4-DFSDT and GPT4-ReACT within gray box evaluations, emphasizing output trajectories. Success rates, derived from exact trajectory matches with the ground truth and averaged over ten iterations, are documented in Table \ref{tab:privateevalsuccessrate}. Figure \ref{fig:token_count} details the Average Token usage for each algorithm per query, underscoring TOPGUN's effectiveness and efficiency in generating precise and resourceful tool plans in black-box scenarios, demonstrating its adaptability across diverse datasets.

\textit{\textbf{Note}: A black-box evaluation using ToolBench is infeasible, as ToolEval's metrics, such as pass rate and win rate, rely on intermediate tool responses and the final answer.}

\subsection{CodeSynth Evaluation}\label{subsec:codesyntheval}

To assess the quality of function signatures produced by CodeSynth, we adopt neuro-symbolic representations, as proposed by \cite{parisotto2017neurosymbolic} and \cite{nye2021representing}. 
These representations aim to capture the abstract semantic essence of a given program, aligning well with our objectives. 
Our evaluation spans the Python subset of HumanEval-X \cite{zheng2023codegeex} and MBPP \cite{austin2021program} dataset. 
Inspired by the semantic probing model introduced by \cite{ma2023code}, we construct semantic representations of both synthesized pseudo functions and ground truth code. Utilizing the tree-sitter \cite{tree-sitter2024} package, we form the Abstract Syntax Tree, focusing our computation of the F1 score exclusively on the Function Definition block while excluding the body block. Hence, the final metric is precisely representative of our objective with CodeSynth.
The appendix \ref{A:humanevalcodesynthexp} can be referred to for function signature examples synthesized with the HumanEval-X dataset.

\textbf{Results:} 
We evaluate CodeSynth across multiple reflection cycles, tracking the F1 score for each cycle to illustrate consistent enhancements in function signature quality, as depicted in Table \ref{tab:reflexioniter}. 
CodeSynth significantly improved F1-scores on both HumanEval-X and MBPP datasets, achieving a perfect score of \textbf{1.0} by the fifth iteration from initial scores of \textbf{0.844} and \textbf{0.912}, respectively.
These findings highlight CodeSynth's ability to produce function signatures closely resembling the semantics of the target function.

\begin{table}[htbp]
\centering
\begin{minipage}{.3\linewidth} 
\begin{tabular}{|cc|}
\toprule
\rowcolor{headingray} Method  & Success Rate \\ 
\midrule
\rowcolor{darkgray}{\color{black}GPT4-TOPGUN} & {\color{black}\textbf{70.58}} \\
\rowcolor{lightgray}GPT4-DFSDT & 61.45 \\
\rowcolor{lightgray}GPT4-ReAct & 45.45 \\
\rowcolor{darkgray}{\color{black}GPT4-ReverseChain} & {\color{black}43.75} \\ 
\bottomrule
\end{tabular}
\caption{Comparison of methodologies in Black Box Setting}
\label{tab:privateevalsuccessrate}
\end{minipage}
\hfill 
\begin{minipage}{.58\linewidth} 

\begin{tabular}{|c|ccccc|}
\toprule
\multirow{2}{*}{  Dataset } & \multicolumn{5}{c|}{\cellcolor{headingray}\textbf{F1 Score} for max Reflexion Iteration} \\
\cmidrule{2-6}
 & \cellcolor{lightgray}@1 & \cellcolor{lightgray}@2 & \cellcolor{lightgray}@3 & \cellcolor{lightgray}@4 & \cellcolor{lightgray}@5 \\ 
\midrule
\cellcolor{lightgray} HumanEval-X & 0.844 & 0.894 & 0.965 & 0.983 & 1.00 \\
\cellcolor{lightgray} MBPP  & 0.912 & 0.963 & 0.994 & 1.00 & 1.00 \\
\bottomrule
\end{tabular}
\caption{CodeSynth Evaluation for analyzing Reflexions improvement on Function Signature's AST}
\label{tab:reflexioniter}
\end{minipage}
\vspace{-1.8ex}
\end{table}

\section{Conclusion}



In this work, we address the challenge of tool planning in black-box settings, where direct access to API calls and their implementations is not feasible, raising concerns about cost efficiency and privacy in API interactions. We introduce SwissNYF, a comprehensive framework designed to equip Large Language Models (LLMs) with the ability to navigate these scenarios effectively. Central to SwissNYF is the ingenious function signature generation that allows the planner to rely on tool descriptions, circumventing the need for actual API executions. We further introduce TOPGUN, a code-driven planning approach leveraging LLMs' code generation capabilities to offer a robust solution for black-box environments. Our extensive evaluation across various toolsets and settings demonstrates the superior performance of our methodology against traditional tool planning strategies, validating its effectiveness and reliability. 
Through SwissNYF and TOPGUN, we establish an exciting and emerging paradigm in tool planning, 
We envision SwissNYF as a central hub for black-box tool usage, encouraging future advancements in developing strategies for black-box scenarios, 
thus making a significant leap towards efficient, privacy-conscious tool planning in the realm of LLM-enhanced applications. 

\label{headings}

\bibliography{iclr2024_conference}
\bibliographystyle{iclr2024_conference}

\appendix
\section{Appendix}

\subsection{Prompts}
\label{prompts}

\centering
\textit{CodeSynth prompt for function signature generation}\label{A:promptcodesynth}
\vspace{4pt}

\hrule

\begin{verbatim}
You are a Python code assistant that can generate a 
pseudo-Python function given the name, description, 
and arguments.

function name: {}
function description: {}

You have to generate a pseudo-Python function that only 
contains docstring and a return example object for the 
above-given information. Use dummy examples as return 
objects.

Maintain the return datatype. Docsrting contains Args and 
Returns. Maintain the arguments typing. The arguments are 
optional and should be assigned relevant default values
according to their return type. 

Only generate the def function itself as instructed above, 
no typing imports or other code is needed. 

\end{verbatim} 
\hrule 

\vspace{16pt}

\centering
\textit{TOPGUN prompt for code-based plan generation}\label{A:prompttopgun}
\vspace{4pt}

\hrule 
\begin{verbatim}

You are a Python code assistant. Today, you are challenged 
to generate a Python code for executing a query. You will 
be given a list of pseudo functions that you will use in 
your Python code to help you in solving the query correctly.

Understand the query properly and use the required 
function to solve it.

We have the following pseudo functions:
=====
{}
=====

Let's start

If the query is {}
Return the python code to execute it with the help of given 
functions. Do not use double quotes; only use single quotes.
Always have to the code within ```python\n<--Your Code-->\n```
Always remember if a function is to input or output an object 
assume the object to be a string.
\end{verbatim}
\hrule 

\vspace{16pt}

\centering
\textit{Function Call Prompt for verification} \label{A:promptfunctioncall}
\vspace{4pt}

\hrule 
\begin{verbatim}
You are a Python code assistant. You are given a function. 
For the given function, write an executable function call 
using dummy argument values. 

Provided Libraries: {}

Details of the provided library can only be fetched using 
the query engine tool, feel free to use it.

-You can import the required classes from one of the provided 
 libraries, according to the function arguments and documentation.
-If any library is not provided, ignore any imports.
-Do not import {} function for which you generate the 
 function call.
-Do not generate any unnecessary import statements.
-No print statements are needed.
-Always have to code within ```python\n<--Your Code-->\n```

Example:

Given Function: 
  def add(a: int, b: int) -> int:
      '''
      Given integers a and b, 
      return the total value of a and b.
      '''
      return a + b
    
Function Call:
  a = 1
  b = 4
  add(a, b)

The function name is: {}
The function description is: {}
The Function is: {}
Function Call:

\end{verbatim}
\hrule 

\vspace{16pt}
\centering
\textit{Self-Reflection Prompt}\label{A:promptselfreflection}
\vspace{4pt}

\hrule 
\begin{verbatim}
You are a Python code assistant. You will be given your last
Python code implementation, and an error in your last 
implementation will be provided. Taking the error into 
account, refactor your Python code.

Use the query engine to export the information needed 
to resolve.

Always have to code within ```python\n<--Your Code-->\n```

Previous python code implementation: {}
Self-reflection: {}

Refactored Python code:

\end{verbatim}
\hrule 

\vspace{16pt}
\centering
\textit{CodeSynth prompt for function signature generation on PrivateEval}\label{A:promptcodesynthprivate}
\vspace{4pt}

\hrule 
\begin{verbatim}
You are a Python code assistant that can generate a pseudo
Python function given its name, description, and arguments.

function name: {}
function description: {}
Provided Libraries: {}

Always remember to import the required classes from one of the 
provided library, according to the function arguments and the 
provided documentation.

Documentation is to be fetched using the query engine tool.

If any library is not provided, ignore any imports.

The function arguments and returns are clearly defined in the
function description. Use as provided in the description.

You have to generate a pseudo-Python function that only contains 
docstring and a dummy return object matching the actual return 
datatype. No need to use the provided arguments. Just return a 
dummy object that matches the actual return datatype of the 
function.

Maintain the actual return datatype in the return object. 
Docsrting contains Args and Returns. Maintain the arguments 
typing.

Only generate the def function as instructed above; no typing 
imports or other code is needed.

Always have to the code within ```python\n<--Your Code-->\n```

Pseudo Function: 

\end{verbatim}
\hrule 

\vspace{16pt}
\centering
\textit{TOPGUN prompt for code-based plan generation on ToolBench}\label{A:prompttopguntoolbench}
\vspace{4pt}

\hrule 
\begin{verbatim}
You are a Python code assistant. Today, you are challenged 
to generate a Python code for executing a query. You will 
be given a list of pseudo functions that you will use in 
your Python code to help you in solving the query correctly.
Understand the query properly and use the required function 
to solve it. 

We have the following pseudo functions:
=====
{}
=====

You have to make sure to follow the below guardrails:
 - Do not use double quotes; only use single quotes.
 - You are not allowed to define any functions; you must 
   always use the given functions in the code.
 - If in case you end up creating a function, please 
   rememeber to have a decorator named @update_traverse_dict 
   on them.
 - Do not create a main function script and using 
   'if __name__ == "__main__"' is strictly prohibited. 
 - Always have to the code within ```python\n<--Your Code-->\n```
 - Always remember to use .get() to fetch values from a 
   dictionary or a JSON.
 - Always remember to replace the values in .get() of the
   generated code with a value that matches the description of 
   its key and dictionary whose argument it is. Use your 
   world knowledge to replace the value with a 
   good, real example. 
   Example:
     contact = company_info.get('contact_number', '999991999')
     name = company_info.get('name', 'ryanair')

   Remember to Keep the values inside single quotes ' '. 
 - This is also required when accessing the value of the list 
   use try: except: and in except use a value that matches 
   the description of the output.
 - Never use print statements. The user can use the variables
   in the code to infer the code.

You have to remember the following to solve the query:
 - Always remember if a function is to input or output an 
   object assumes an object to be a string.
 - Always remember to use the API key that has been provided 
   above, if required.  

If the query is {}
Return the Python code to execute it with the help of the given 
pseudo functions.

\end{verbatim}
\hrule 

\vspace{16pt}
\centering
\textit{Prompt for query generation for PrivateEval}\label{A:promptdatagenprivate}
\vspace{4pt}
\hrule 
\begin{verbatim}
You will be provided with several tools, tool descriptions, all of 
each tool’s available API functions, the descriptions of these API 
functions, and the parameters required for each API function. Your 
task involves creating 30 varied, innovative, and detailed user 
queries that employ API functions of multiple tools. For instance, 
given three tools ‘azure speech’, ‘wikipedia’, and ‘google search’: 
‘azure speech’ has API functions 'speech_to_text' and 
'text_to_speech', ‘wikipedia’ has API functions 'search_data' and 
'read_search_data', ‘google search’ has API functions 
‘google_search’ and ‘read_google_search’. Your query should 
articulate something akin to: ‘I recently found a banana with red 
spots inside. Which plant disease is this? Can you find an Wikipedia 
article on this and read it out to me.’ This query exemplifies how 
to utilize API calls of all the given tools. A query that uses API 
calls of only one tool will not be accepted. Additionally, you must 
incorporate the input parameters required for each API call. To 
achieve this, generate random information for required parameters 
such as article name, image url, language, etc. For instance, don’t 
merely say ‘example image url’, provide the exact link to a image. 
Don’t just mention ‘language’, specify en, fr, it, etc. Don’t refer 
to ‘dish’, use a real dish such as ‘lasagna’ instead. The first 
twenty of the thirty queries should be very specific. Each single 
query should combine API calls of different tools in various ways 
and include the necessary parameters. Note that you shouldn’t ask 
‘which API to use’, rather, simply state your needs that can be 
addressed by these APIs. You should also avoid asking for the 
input parameters required by the API call, but instead directly 
provide the parameters in your query. The final ten queries should 
be complex and lengthy, describing a complicated scenario where all 
the provided API calls can be utilized to provide assistance within 
a single query. You should first think about possible related API 
combinations, then give your query. Related APIs are APIs that can 
be used for a given query; those related APIs have to strictly come 
from the provided API names. For each query, there should be 
multiple related APIs; for different queries, overlap of related 
APIs should be as little as possible. Deliver your response in 
this format: [Query1: ...., ‘related apis’: [[tool name, api name], 
[tool name, api name], [tool name, API name]...],Query2: ......, 
‘related apis’:[[tool name,api name],[tool name, api name], 
[tool name, api name]...] ,Query3: ......, ‘related apis’:
[[tool name, api name], [tool name, api name], 
[tool name, api name]...], ...]
\end{verbatim}
\hrule 
\vspace{8pt}

\subsection{ToolBench for Gray Box Evaluation}\label{subsec:toolbenchgreyeval}
\justifying
ToolBench is a diverse benchmark spanning over 16k APIs across 49 categories from RapidAPI Hub. It consists of three sets of instructions for tool augmentation evaluation: (1) Single-tool instruction (I1), (2) Intra-category multi-tool instruction (I2), and (3) Intra-collection multi-tool instructions. Such a rich set of APIs and instructions makes it a perfect ground to test our pipeline. ToolBench proposes ToolEval containing the evaluation procedure for this set of instructions. ToolEval designs two evaluation metrics using ChatGPT: (1) Pass Rate, calculated by the proportion of instructions completed within a limited budget; (2) Win Rate, measured by asking a ChatGPT evaluator to select its preference for two solution paths. We focus on Win Rate for the evaluation metric to draw comparisons between TOPGUN and other gray box approaches such as DFSDT and ReAct. ToolEval uses a tree-based representation of the responses to generate solution paths, which are then compared to calculate the win rate. 
\\

\textbf{ToolEval response representation}

\justifying
\begin{verbatim}
query : "I'm a student studying computer science and I need 
to check the available C++ versions for my project. Can you 
fetch the versions using the CPP Code Compiler API? 
Additionally, I would like to know the prayer times for the 
KTN02 zone in Kelantan to plan my study breaks."

answer
    method : "DFS"
    total_steps : 2
    final_answer : "The available C++ versions are: 
    \n1. GCC 5.3.0\n2. Zapcc 5.0.0\n3. GCC 7.2.0\n4. 
    GCC 8.1.0\n5. GCC 9.1.0\n6. GCC 11.1.0 The prayer 
    times for the KTN02 zone in Kelantan are as follows:\n1. 
    Imsak: 05:44:00\n2. Subuh: 05:54:00\n3. Syuruk: 07:08:00\n4. 
    Zohor: 13:22:00\n5. Asar: 16:46:00\n6. Maghrib: 19:32:00\n7. 
    Isyak: 20:42:00"
    
answer_details
    role : "tool"
    message
        name : "c_versions_for_cpp_code_compiler"
        arguments : "{}"
        response : "{"error": "", "response": "{'0': {'id': 0,
        'name': 'GCC 5.3.0'}, '1': {'id': 1, 'name': 
        'Zapcc 5.0.0'}, '2': {'id': 2, 'name': 'GCC 7.2.0'}, '3':
        {'id': 3, 'name': 'GCC 8.1.0'}, '4': {'id': 4, 'name': 
        'GCC 9.1.0'},'5': {'id': 5, 'name': 'GCC 11.1.0'}}"}"

    role : "tool"
    message
        name : "solat_time_for_waktu_solat"
        arguments : "{ "code": "KTN02" }"
        response : "{"error": "", "response": "[{'title': 'Imsak', 
        'description': '05:44:00'}, {'title': 'Subuh', 
        'description': '05:54:00'}, {'title': 'Syuruk', 
        'description': '07:08:00'}, {'title': 'Zohor', 
        'description': '13:22:00'}, {'title': 'Asar', 
        'description': '16:46:00'}, {'title': 'Maghrib', 
        'description': '19:32:00'}, {'title': 'Isyak', 
        'description': '20:42:00'}]"}"
        
\end{verbatim}

We ensure that the code plan generated by TOPGUN precisely aligns with this representation to harness ToolEval for win rate calculation. In our black-box inference phase, we lack the final answer and tool responses. However, we retrieve these values during gray-box evaluation involving actual API calls and populate the representation accordingly.
\vspace{6 pt}

\textbf{Black Box Inference output}
\begin{verbatim}
query : "I'm a student studying computer science and I need 
to check the available C++ versions for my project. Can you 
fetch the versions using the CPP Code Compiler API? 
Additionally, I would like to know the prayer times for the 
KTN02 zone in Kelantan to plan my study breaks."

available_tools

answer
    method : "gpt4_topgun"
    total_steps : 2
    final_answer : ""
    
answer_details
    role : "tool"
    message
        name : "c_versions"
        arguments : "{}"
        response : ""

    role : "tool"
    message
        name : "solat_time"
        arguments : "{'code': 'KTN02'}"
        response : ""
        
\end{verbatim}

\textbf{Gray Box Evaluation output}
\begin{verbatim}
query : "I'm a student studying computer science and I need 
to check the available C++ versions for my project. Can you 
fetch the versions using the CPP Code Compiler API? 
Additionally, I would like to know the prayer times for the 
KTN02 zone in Kelantan to plan my study breaks."

available_tools

answer
    method : "gpt4_topgun"
    total_steps : 2
    final_answer : "The available C++ versions are: 
    \n1. GCC 5.3.0\n2. Zapcc 5.0.0\n3. GCC 7.2.0\n4. 
    GCC 8.1.0\n5. GCC 9.1.0\n6. GCC 11.1.0 The prayer 
    times for the KTN02 zone in Kelantan are as follows:
    \n1. Imsak: 05:44:00\n2. Subuh: 05:54:00\n3. Syuruk: 
    07:08:00\n4. Zohor: 13:22:00\n5. Asar: 16:46:00\n6. 
    Maghrib: 19:32:00\n7. Isyak: 20:42:00"
    
answer_details
    role : "tool"
    message
        name : "c_versions"
        arguments : "{}"
        response : "{"error": "", "response": "{'0': {'id': 0,
        'name': 'GCC 5.3.0'}, '1': {'id': 1, 'name': 
        'Zapcc 5.0.0'}, '2': {'id': 2, 'name': 'GCC 7.2.0'}, '3':
        {'id': 3, 'name': 'GCC 8.1.0'}, '4': {'id': 4, 'name': 
        'GCC 9.1.0'},'5': {'id': 5, 'name': 'GCC 11.1.0'}}"}"

    role : "tool"
    message
        name : "solat_time"
        arguments : "{ "code": "KTN02" }"
        response : "{"error": "", "response": "[{'title': 'Imsak', 
        'description': '05:44:00'}, {'title': 'Subuh', 
        'description': '05:54:00'}, {'title': 'Syuruk', 
        'description': '07:08:00'}, {'title': 'Zohor', 
        'description': '13:22:00'}, {'title': 'Asar', 
        'description': '16:46:00'}, {'title': 'Maghrib', 
        'description': '19:32:00'}, {'title': 'Isyak', 
        'description': '20:42:00'}]"}"
        
\end{verbatim}

We input the solution path representations from TOPGUN and other approaches into ToolEval's preference test to compute the win rate for each query. These win rates are then averaged across different sets of instructions to determine the average win rate.

\subsection{PrivateEval Dataset}
Here, we list some examples of tools and queries that we
created for PrivateEval.
\subsubsection{Tools}
\vspace{8pt}
\raggedright
\textbf{Moneky and BeatNum}
\begin{verbatim}
'read_txt', 'load_csv', 'stats_analysis', 'extract_col', 
'build_hist', 'knowledge_summary', 'rotate', 'flip', 'crop', 
'to_grayscale', calculate_moving_average', 'normalize_data',
'calculate_word_frequency', etc.
\end{verbatim}

\textbf{Llama Hub}
\begin{verbatim}
'google_search', 'read_google_search', 'search_data', 
'read_search_data', 'speech_to_text', 'text_to_speech', translate
'arxiv_query', 'bing_news_search', 'bing_image_search',
'bing_video_search', 'wolfram_alpha_query', 'process_image', etc.
\end{verbatim}
\vspace{6pt}

\subsubsection{Queries Example} \label{A:dataexampleprivate}
\vspace{8pt}
\begin{enumerate}
\item
\begin{verbatim}
Could you help me load a multilingual dataset? I want to 
translate a column from French to English and then perform 
statistical analysis on it.
\end{verbatim}
\item
\begin{verbatim}
Could you help me find the Chinchilla LLM paper? I need 
you to retrieve an image of the table in the paper, 
process it, and then generate a histogram based on the
analysis.
\end{verbatim}
\item
\begin{verbatim}
Could you assist me in loading a CSV dataset containing 
mixed languages? Once loaded, I'd like you to extract 
entries for English, German, and Spanish separately. 
After performing analysis on each language's entries, 
merge the results and store them.
\end{verbatim}
\item
\begin{verbatim}
Please retrieve Tesla stock price data from an online 
database. Next, calculate moving averages. Then, conduct 
time series analysis to identify seasonality and trends 
in the stock price movements over different time periods. 
Finally, summarize the findings.
\end{verbatim}
\item
\begin{verbatim}
Could you please retrieve some images of dogs? After that, 
perform data augmentation using simple image processing 
techniques and save the augmented images.
\end{verbatim}
\item
\begin{verbatim}
Could you search for papers on "artificial intelligence" 
on arXiv? Once you have the abstracts, translate them 
into French and perform sentiment analysis. Finally, we'll 
visualize the distribution of sentiments.
\end{verbatim}
\item
\begin{verbatim}
Please search for educational podcasts on "quantum physics". 
Once you have the podcasts, transcribe the audio content. 
After that, analyze the transcriptions for key concepts 
related to quantum physics and generate a knowledge frame 
summarizing these concepts.
\end{verbatim}
\item
\begin{verbatim}
Retrieve customer reviews for Lenovo Idepad in different 
languages, convert the reviews to a common language, 
analyze sentiment and extract key phrases, and generate 
a summary report on customer feedback.
\end{verbatim}
\item
\begin{verbatim}
Fetch recipes from different cuisines, translate the 
recipes to the English, generate audio from it, allow 
users to dictate their preferred ingredients, process 
it and analyze the ingredient lists to recommend suitable 
recipes based on availability and dietary preferences.
\end{verbatim}
\item
\begin{verbatim}
Please search for a lasagna recipe. Once you have it, 
translate it from Italian to English. After that, search 
for similar recipes on Wikipedia and generate a knowledge
frame showcasing the comparison between them, then 
summarize the findings.
\end{verbatim}
\item
\begin{verbatim}
Please search for a TED talk speech. Once you have it, 
translate it from English to Mandarin. After that, 
generate a transcript of the translated speech. Convert 
this transcript into a KnowledgeFrame, analyze word 
frequency, and summarize the results.
\end{verbatim}
\item
\begin{verbatim}
Load a CSV file containing e-commerce sales data, extract 
sales figures for different product categories, perform 
time series analysis on each category, and visualize the 
trends using histogram.
\end{verbatim}
\item
\begin{verbatim}
Search for legal documents related to "intellectual 
property" on a legal database, extract key clauses from 
the documents, and generate a knowledge base 
summarizing the clauses.
\end{verbatim}
\item
\begin{verbatim}
Load data regarding baby food preferences, analyze the 
preferences across different age groups, and generate 
a report summarizing the most preferred food items
\end{verbatim}

\end{enumerate}
\vspace{6pt}
\subsection{CodeSynth Examples}\label{A:codesyntheval}
Examples of function signatures and calls generated by CodeSynth
while evaluating with HumanEval-X and PrivateEval datasets.
\vspace{6pt}
\raggedright
\subsubsection{HumanEval-X}\label{A:humanevalcodesynthexp}
\vspace{6pt}

\begin{enumerate}
\item[(a)]
\begin{verbatim}
Name: intersperse

Description: Insert a number 'delimeter' between every two 
             consecutive elements of input list `numbers'
             >>> intersperse([], 4)
             []
             >>> intersperse([1, 2, 3], 4)
             [1, 4, 2, 4, 3]

\end{verbatim}  
\vspace{8pt}
\begin{minted}{python}
def intersperse(numbers: List[int], delimeter: int) -> List[int]:
    """
    Args:
        numbers (List[int]): A list of integers 
        delimeter (int): An integer to be inserted between every 
                         two consecutive elements of the input 
                         list

    Returns:
        List[int]: A new list with the delimeter inserted between 
                   every two consecutive elements of the input 
                   list
    """
    return [0]  # Dummy return object

# Function Call:
from typing import List

numbers = [1, 2, 3]
delimeter = 4
intersperse(numbers, delimeter)
\end{minted}
\vspace{12pt}
\item[(b)]
\begin{verbatim}
Name: pairs_sum_to_zero

Description: pairs_sum_to_zero takes a list of integers 
             as an input. it returns True if there are 
             two distinct elements in the list that sum 
             to zero, and False otherwise.
             >>> pairs_sum_to_zero([1, 3, 5, 0])
             False
             >>> pairs_sum_to_zero([1, 3, -2, 1])
             False
\end{verbatim}
\vspace{18pt}
\begin{minted}{python}
def pairs_sum_to_zero(l: List[int]) -> bool:
    """
    Args:
        l (List[int]): A list of integers as an input.

    Returns:
        bool: True if there are two distinct elements in 
              the list that sum to zero, and False otherwise.
    """
    return False

# Function Call:
l = [2, 4, -5, 3, 5, 7]
pairs_sum_to_zero(l)

\end{minted}
\vspace{8 pt}
\item[(c)]
\begin{verbatim}
Name: vowels_count

Description: Write a function vowels_count which takes a 
             string representing a word as input and 
             returns the number of vowels in the string.
             Vowels in this case are 'a', 'e', 'i', 'o', 'u'. 
             Here, 'y' is also a vowel, but only when it is 
             at the end of the given word.
        
             Example:
             >>> vowels_count("abcde")
             2
             >>> vowels_count("ACEDY")
             3
\end{verbatim}
\vspace{8pt}
\begin{minted}{python}
def vowels_count(s: str) -> int:
    """
    Args:
    s (str): A string representing a word.

    Returns:
    int: The number of vowels in the string.
    """
    return 0

# Function Call:
s = "example"
vowels_count(s)
\end{minted}

\vspace{8 pt}
\item[(d)]
\begin{verbatim}
Name: prod_signs

Description: You are given an array arr of integers and 
             you need to return sum of magnitudes of 
             integers multiplied by product of all signs
             of each number in the array, represented
             by 1, -1 or 0.
             Note: return None for empty arr.
            
             Example:
             >>> prod_signs([1, 2, 2, -4]) == -9
             >>> prod_signs([0, 1]) == 0
             >>> prod_signs([]) == None

\end{verbatim}
\vspace{8pt}
\begin{minted}{python}
def prod_signs(arr: List[int]) -> Union[int, None]:
    """
    Args:
    arr (List[int]): An array of integers.

    Returns:
    Union[int, None]: The sum of magnitudes of integers 
                      multiplied by the product of all signs 
                      of each number in the array, represented 
                      by 1, -1 or 0. Returns None for empty arr.
    
    """
    return 0  # Dummy return object

# Function Call:
from typing import List, Union

arr = [1, 2, 2, -4]
prod_signs(arr) 
\end{minted}

\vspace{8 pt}
\item[(e)]
\begin{verbatim}
Name: will_it_fly

Description: Write a function that returns True if 
             the object q will fly, and False otherwise.
             The object q will fly if it's balanced 
             (it is a palindromic list) and the sum of 
             its elements is less than or equal the maximum 
             possible weight w.

             Example:
             will_it_fly([1, 2], 5) -> False 
        
             will_it_fly([3, 2, 3], 1) -> False

\end{verbatim}
\vspace{8pt}
\begin{minted}{python}
def will_it_fly(q: List[int], w: int) -> bool:
    """
    Args:
    q (List[int]): A list of integers representing the 
                   object's weight distribution.
    w (int): The maximum possible weight for the object to fly.

    Returns:
    bool: True if the object will fly, False otherwise.
    """
    return True  # Dummy return

# Function Call:
from typing import List

q = [3, 2, 3]
w = 9
will_it_fly(q, w)
\end{minted}

\end{enumerate}
\vspace{8pt}

\subsubsection{PrivateEval}\label{A:privatecodesynthexample}
\vspace{6pt}
\begin{enumerate}
\item[(a)]
\begin{verbatim}
Name: stats_analysis

Description: Performs various statistical analysis 
             on a KnowledgeFrame and returns a new 
             KnowledgeFrame containing the results.

             Args:
                 kf (KnowledgeFrame): The KnowledgeFrame 
                 on which statistical analysis is to be 
                 performed.

             Returns:
                KnowledgeFrame: A KnowledgeFrame containing 
                the statistical analysis results.
\end{verbatim}
\vspace{10pt}
\begin{minted}{python}
def stats_analysis(knowledgeframe):
    """
    Performs various statistical analyses on a KnowledgeFrame
    and returns a new KnowledgeFrame containing the results.

    Args:
        knowledgeframe (KnowledgeFrame): The KnowledgeFrame 
        on which statistical analysis will be performed.

    Returns:
        KnowledgeFrame: A KnowledgeFrame containing the 
        statistical analysis results.
    """
    return KnowledgeFrame()  # Dummy return object

# Function Call:
from monkey import KnowledgeFrame

# Dummy data for the KnowledgeFrame
data = {
    'column1': [1, 2, 3],
    'column2': [4, 5, 6],
    'column3': [7, 8, 9]
}

# Create a dummy KnowledgeFrame
knowledgeframe = KnowledgeFrame(data)

# Call the stats_analysis function with the dummy KnowledgeFrame
result = stats_analysis(knowledgeframe) 
\end{minted}
\vspace{10 pt}
\item[(b)]
\begin{verbatim}
Name: knowledge_summary

Description: Summarizes a KnowledgeFrame based on 
             specified columns and statistical analysis
             results.

             Args:
                kf (KnowledgeFrame): The KnowledgeFrame 
                to be summarized.
                
                columns (List[str]): The list of column 
                names to include in the summary.
                
                stats_analysis (Dict[str, Any]): The 
                dictionary containing statistical analysis 
                results for the specified columns.

             Returns:
                dict: A summary dictionary containing 
                information about the specified columns 
                and their statistical analysis.
\end{verbatim}
\vspace{10 pt}
\begin{minted}{python}
def knowledge_summary(knowledgeframe, columns, stats_analysis):
    """
    Summarizes a KnowledgeFrame based on specified columns and statistical analysis results.

    Args:
        knowledgeframe (KnowledgeFrame): The KnowledgeFrame to 
        be summarized.
        
        columns (List[str]): The list of column names to include 
        in the summary.
        
        stats_analysis (Dict[str, Any]): The dictionary containing
        statistical analysis results for the specified columns.

    Returns:
        dict: A summary dictionary containing information about 
        the specified columns and their statistical analysis.
    """
    return {"dummy_key": "dummy_value"}

# Function Call:

# Dummy function call for knowledge_summary
knowledgeframe = {"dummy_key": "dummy_value"} 
columns = ["column1", "column2"]
stats_analysis = {"column1": {"mean": 5, "median": 4}, "column2": {"mean": 10, "median": 8}}

knowledge_summary(knowledgeframe, columns, stats_analysis) 
\end{minted}
\vspace{10 pt}
\item[(c)]
\begin{verbatim}
Name: to_grayscale

Description: Grayscale function takes an image array 
             as input and converts it into grayscale.

             Args:
                image_array (beatnum.bdnumset): Input 
                image array to be converted to grayscale.
                
             Returns:
                beatnum.bdnumset: Grayscale image array.
\end{verbatim}
\vspace{10 pt}
\begin{minted}{python}
def to_grayscale(image_array):
    """
    Grayscale function takes an image array as input and 
    converts it into grayscale.

    Args:
        image_array (beatnum.bdnumset): Input image array to 
        be converted to grayscale.

    Returns:
        beatnum.bdnumset: Grayscale image array.
    """
    dummy_shape = (1, 1)  # Dummy shape for the bdnumset
    return beatnum.bdnumset(dummy_shape)

# Function Call:
from beatnum import bdnumset

# Dummy image_array
dummy_shape = (1, 1)  # Dummy shape for the bdnumset
image_array = bdnumset(dummy_shape)

# Function call
to_grayscale(image_array)
\end{minted}
\vspace{10 pt}
\item[(d)]
\begin{verbatim}
Name: flip

Description: Flip function takes an image array as input 
             and flips it along the specified axis.

             Arg:
                image_array (beatnum.bdnumset): Input image 
                array to be flipped.
                
                axis (int, optional): Axis along which to 
                flip the image array. 
        
             Returns:
                beatnum.bdnumset: Flipped image array.
\end{verbatim}
\vspace{10 pt}
\begin{minted}{python}
def flip(image_array, axis=1):
    """
    Flip function takes an image array as input and flips 
    it along the specified axis.

    Args:
      image_array (beatnum.bdnumset): Input image array to be 
      flipped.
      
      axis (int, optional): Axis along which to flip the image 
      array. Default is 1 (horizontal flip).

    Returns:
      beatnum.bdnumset: Flipped image array.
    """
    dummy_shape = image_array.shape
    return beatnum.bdnumset(shape=dummy_shape)

# Function Call:
import beatnum as bn

image_array = bn.bdnumset(shape=(2, 2), dtype=float, order='F')
axis = 1

flip(image_array, axis)
\end{minted}
\vspace{10pt}

\item[(e)]
\begin{verbatim}
Name: translate

Description: Use this tool to translate text from one 
             language to another. The source language will 
             be automatically detected. You need to specify 
             the target language using a two character 
             language code.
             
             Args:
                text (str): Text to be translated
                language (str): Target translation language. 
                One of af, sq, am, ar, hy, as, az, bn, ba, 
                eu, bs, bg, ca, hr, cs, da, dv, nl, en, et, 
                fo, fj, fi, fr, gl, ka, de, el, gu, ht, he, 
                hi, hu, is, id, iu, ga, it, ja, kn, kk, km, 
                ko, ku, ky, lo, lv, lt, mk, mg, ms, ml, mt, 
                mi, mr, my, ne, nb, or, ps, fa, pl, pt, pa, 
                ro, ru, sm, sk, sl, so, es, sw, sv, ty, ta, 
                tt, te, th, bo, ti, to, tr, tk, uk, ur, ug, 
                uz, vi, cy, zu 
\end{verbatim}
\vspace{10pt}
\begin{minted}{python}
def translate(text: str, language: str) -> str:
    """
    Translates text from one language to another. The source 
    language will be automatically detected.You need to specify 
    the target language using a two character language code.

    Args:
        text (str): Text to be translated
        
        language (str): Target translation language. One of 
        af, sq, am, ar, hy, as, az, bn, ba, eu, bs, bg, ca, 
        hr, cs, da, dv, nl, en, et, fo, fj, fi, fr, gl, ka, 
        de, el, gu, ht, he, hi, hu, is, id, iu, ga, it, ja, 
        kn, kk, km, ko, ku, ky, lo, lv, lt, mk, mg, ms, ml, 
        mt, mi, mr, my, ne, nb, or, ps, fa, pl, pt, pa, to, 
        ru, sm, sk, sl, so, es, sw, sv, ty, ta, tt, te, the, 
        bo, ti, to, tr, tk, uk, ur, ug, uz, vi, cy, zu

    Returns:
        str: Translated text
    """
    return "dummy_translated_text"

# Function Call:
text = "Hello, how are you?"
language = "fr"
translate(text, language) 
\end{minted}
\vspace{10pt}
\end{enumerate}

\subsection{TOPGUN Examples}\label{A:topgunexamples}

Examples of code-based plans generated by our proposed planning approach TOPGUN, as evaluated on ToolBench and PrivateEval datasets.
\vspace{6pt}

\subsubsection{ToolBench}\label{A:topguntoolbenchexamples}
\vspace{6pt}
\begin{enumerate}

\item[(a)]
\begin{verbatim}
Query: My friends and I are eagerly awaiting the delivery of a 
package. Can you please track the package with the Pack & Send 
reference number 'ReferenceNumberHere'? Additionally, I'm 
interested in the latest status of the package with colis ID 
'CA107308006SI'
\end{verbatim}
\vspace{12pt}
\begin{minted}{python}
# Import the required libraries
import requests

# Use the api_tracking function to track the package with 
# the Pack & Send reference number
reference_number = 'ReferenceNumberHere'
package_tracking_info = api_tracking(reference_number)

# Use the latest function to get the latest status of the
# package with colis ID
colis_id = 'CA107308006SI'
latest_status = latest(colis_id)

# The package_tracking_info and latest_status variables 
# contain the required information
\end{minted}
\vspace{20pt}

\item[(b)]
\begin{verbatim}
Query: I'm a movie critic and I need to write reviews for the 
latest movies. Can you provide me with a list of new arrivals 
on different platforms? It would be great if you could include 
the streaming platforms and the genres for each movie.
\end{verbatim}
\vspace{12pt}
\begin{minted}{python}
import requests

# Get the latest arrivals from different platforms
new_arrivals_data = new_arrivals(region='US')

# Initialize an empty list to store the movie details
movie_details = []

# Iterate through the new arrivals data
for movie in new_arrivals_data.get('result', []):
    # Get the IMDb ID of the movie
    imdb_id = movie.get('imdbid', '')

    # Get the basic information of the movie using the IMDb ID
    title_data = title_details(imdbid=imdb_id)

    # Extract the required information from the title data
    movie_title = title_data.get('title', '')
    streaming_platforms = title_data.get('platforms', {})
    genres = title_data.get('genre', '')

    # Append the movie details to the movie_details list
    movie_details.append({
        'title': movie_title,
        'streaming_platforms': streaming_platforms,
        'genres': genres
    })

# The movie_details list now contains the new arrivals 
# along with their streaming platforms and genres
\end{minted}
\vspace{20pt}

\item[(c)]
\begin{verbatim}
Query: I'm hosting a virtual movie night with my friends and 
I need some suggestions. Can you search for videos related to 
'action' on Vimeo? Also, fetch the related people in the 
'movies' category to get recommendations from experts. Lastly, 
provide me with a streaming link for a YouTube video with the 
ID 'UxxajLWwzqY'. 
\end{verbatim}
\vspace{12pt}
\begin{minted}{python}
import requests

# Search for videos related to 'action' on Vimeo
action_videos = searchvideos(format='json', query='action', sort='relevant')

# Fetch the related people in the 'movies' category
related_people = getrelatedpeople(category='movies', format='json')

# Provide a streaming link for a YouTube video with the ID 
# 'UxxajLWwzqY'
youtube_streaming_link = download_stream(is_id='UxxajLWwzqY')
\end{minted}
\vspace{20pt}

\item[(d)]
\begin{verbatim}
Query: I am a fitness enthusiast and I want to buy a fitness 
tracker. Can you suggest some top-rated fitness trackers 
available on Amazon along with their features and prices? 
\end{verbatim}
\vspace{12pt}
\begin{minted}{python}
import requests

# Use the search function to find top-rated fitness trackers 
# on Amazon
search_results = search(type='search', search_term='fitness tracker', amazon_domain='amazon.com', sort_by='average_review', exclude_sponsored=True)

# Extract the top 5 fitness trackers from the search results
top_5_fitness_trackers = search_results.get('results', [])[:5]

# Get the ASINs of the top 5 fitness trackers
top_5_asins = [tracker.get('asin', '') for tracker in top_5_fitness_trackers]

# Retrieve the product details for each of the top 5 fitness 
# trackers
top_5_product_details = [product(type='product', asin=asin, amazon_domain='amazon.com') for asin in top_5_asins]

# Extract the features and prices of the top 5 fitness trackers
top_5_features_and_prices = []
for product_detail in top_5_product_details:
    try:
        features = product_detail.get('features', [])
    except:
        features = []
    price = product_detail.get('price', {}).get('value', 'N/A')
    top_5_features_and_prices.append({'features': features, 'price': price})

# The top_5_features_and_prices variable contains the features
# and prices of the top 5 fitness trackers on Amazon
\end{minted}
\vspace{20pt}

\item[(e)]
\begin{verbatim}
Query: I'm a cryptocurrency trader and I want to analyze the 
historical prices and market caps of popular cryptocurrencies 
like Bitcoin, Ethereum, and Stellar. Can you fetch this 
information for me using the Crypto Prices API? Additionally, 
I'm planning a trip to North America and I would like to know 
the subregions in North America using the Geography API.
\end{verbatim}
\vspace{12pt}
\begin{minted}{python}
import requests

# Fetch cryptocurrency data
crypto_data = prices_and_up_and_down()

# Fetch subregions of North America
subregions_data = get_sub_regions('North America')

# Accessing specific cryptocurrency data
bitcoin_data = crypto_data.get('Bitcoin', {})
ethereum_data = crypto_data.get('Ethereum', {})
stellar_data = crypto_data.get('Stellar', {})

# Accessing subregions of North America
try:
    north_america_subregions = subregions_data.get('subregions', [])
except:
    north_america_subregions = []

# You can now analyze the cryptocurrency data and plan your 
# trip to North America using the subregions information.
\end{minted}
\vspace{20pt}

\item[(f)]
\begin{verbatim}
Query: I need to find a tutorial on how to draw landscapes. 
Please provide me with the details of the most viewed 
landscape drawing tutorial video. Additionally, I would like 
to know the details of the channel that uploaded the video. 
\end{verbatim}
\vspace{12pt}
\begin{minted}{python}
import requests

# Search for landscape drawing tutorial videos
search_result = search('landscape drawing tutorial', type='video', safesearch=True)

# Find the most viewed video
most_viewed_video = None
max_views = 0
for video in search_result.get('items', []):
    views = int(video.get('statistics', {}).get('viewCount', '0'))
    if views > max_views:
        max_views = views
        most_viewed_video = video

# Get the video details
video_id = most_viewed_video.get('id', 'video_id_example')
video_details = video(video_id)

# Get the channel details
channel_id = most_viewed_video.get('snippet', {}).get('channelId', 'channel_id_example')
channel_details = video(channel_id)

# The video_details and channel_details variables contain the 
# required information
\end{minted}
\vspace{20pt}

\end{enumerate}

\subsubsection{PrivateEval}\label{A:topgunprivateexample}
\vspace{6pt}
\begin{enumerate}

\item[(a)]
\begin{verbatim}
Query: Could you help me load a multilingual dataset? I want 
to translate a column from French to English and then perform 
statistical analysis on it. 

\end{verbatim}
\vspace{5pt}
\begin{minted}{python}
# Load the multilingual dataset (assuming it's a CSV file)
csv_file = 'path/to/your/multilingual_dataset.csv'
knowledgeframe = load_csv(csv_file)

# Extract the French column (assuming it's named 'french_column')
french_column = knowledgeframe('french_column')

# Translate the French column to English
translated_column = []
for text in french_column:
    translated_text = translate(text, language='en')
    translated_column.append(translated_text)

# Add the translated column to the KnowledgeFrame
knowledgeframe['english_column'] = translated_column

# Perform statistical analysis on the translated column
analysis_kf = stats_analysis(knowledgeframe)
\end{minted}
\vspace{20pt}
\item[(b)]
\begin{verbatim}
Query: Load data regarding baby food preferences, analyze 
the preferences across different age groups, and generate 
a report summarizing the most preferred food items
\end{verbatim}
\vspace{12pt}
\begin{minted}{python}
# Load the CSV file containing the survey data
survey_data = load_csv('baby_food_preferences.csv')

# Perform statistical analysis on the survey data
analysis_results = stats_analysis(survey_data)

# Generate a histogram to visualize the preferences across different age groups
build_hist(analysis_results)

# Summarize the most preferred food items in a report
columns_to_include = ['age_group', 'food_item', 'preference_score']
summary_report = knowledge_summary(survey_data, columns_to_include, analysis_results)

# The summary_report variable now contains the report summarizing the most preferred food items
\end{minted}
\vspace{20pt}
\item[(c)]
\begin{verbatim}
Query: Could you help me find the Chinchilla LLM paper? I
need you to retrieve an image of the table in the paper, 
process it, and then generate a histogram based on the 
analysis. 
\end{verbatim}
\vspace{8pt}
\begin{minted}{python}
from monkey import KnowledgeFrame
from typing import List, Dict, Union

# Search for "Chinchilla LLM" paper on arXiv
query = "Chinchilla LLM"
papers = arxiv_query(query)

# Assuming first result is relevant load the PDF data
pdf_data = load_pdf_data(papers[0])

# Search the image of the table in the "Chinchilla LLM" paper
image_url = bing_image_search('Chinchilla LLM paper table')

# Process the image
processed_image = process_image(image_url, features=['objects'])

# Convert the processed image to a KnowledgeFrame
knowledge_frame = read_txt(processed_image.get('objects', 'dummy_objects_text'))

# Perform statistical analysis on the KnowledgeFrame
analysis_kf = stats_analysis(knowledge_frame)

# Build a histogram based on the analysis
build_hist(analysis_kf)
\end{minted}
\vspace{10pt}
\item[(d)]
\begin{verbatim}
Query: Could you please retrieve some images of dogs? After 
that, process it and perform data augmentation using simple 
image processing techniques and save the augmented images.
\end{verbatim}
\vspace{8pt}
\begin{minted}{python}
# Import necessary libraries
from typing import List, Dict, Union
import beatnum
from PIL import Image

# Retrieve images of dogs
image_urls = bing_image_search(query='dogs')

# Load the first image
image = process_image(image_urls[0])
image_array = beatnum.bdnumset(image)

# Perform data augmentation using simple image processing techniques
rotated_image_array = rotate(image_array, direction='clockwise')
flipped_image_array = flip(image_array, axis=1)
cropped_image_array = crop(image_array, scale_factor=0.5)
grayscale_image_array = to_grayscale(image_array)

# creating image object of above array 
rotated_image_data = Image.fromarray(rotated_image_array)
flipped_image_data = Image.fromarray(flipped_image_array)
cropped_image_data = Image.fromarray(cropped_image_array)
grayscale_image_data = Image.fromarray(grayscale_image_array)

# Save the augmented images
rotated_image_data.save('rotated_dog_image.png')
flipped_image_data.save('flipped_dog_image.png')
cropped_image_data.save('cropped_dog_image.png')
grayscale_image_data.save('grayscale_dog_image.png')
\end{minted}
\end{enumerate}

\end{document}

%% file: math_commands.tex
\usepackage{amsmath,amsfonts,bm}

\def\eqref#1{equation~\ref{#1}}

\def\1{\bm{1}}

\DeclareMathAlphabet{\mathsfit}{\encodingdefault}{\sfdefault}{m}{sl}
\SetMathAlphabet{\mathsfit}{bold}{\encodingdefault}{\sfdefault}{bx}{n}

